\documentclass[letterpaper]{article} % DO NOT CHANGE THIS
\usepackage{aaai2026}  % DO NOT CHANGE THIS
\usepackage{times}  % DO NOT CHANGE THIS
\usepackage{helvet}  % DO NOT CHANGE THIS
\usepackage{courier}  % DO NOT CHANGE THIS
\usepackage[hyphens]{url}  % DO NOT CHANGE THIS
\usepackage{graphicx} % DO NOT CHANGE THIS
\usepackage{natbib}  % DO NOT CHANGE THIS AND DO NOT ADD ANY OPTIONS TO IT
\usepackage{caption} % DO NOT CHANGE THIS AND DO NOT ADD ANY OPTIONS TO IT
\usepackage[table]{xcolor}
\usepackage[table]{xcolor}  % 支持表格单元格颜色
\usepackage{amsmath} 
\usepackage{algorithm}
\usepackage{algorithmic}

\usepackage{newfloat}
\usepackage{listings}
\DeclareCaptionStyle{ruled}{labelfont=normalfont,labelsep=colon,strut=off} % DO NOT CHANGE THIS
\floatstyle{ruled}
\newfloat{listing}{tb}{lst}{}
\floatname{listing}{Listing}
\usepackage{amssymb}
 \usepackage{amsmath}
 \usepackage{bm}
 \usepackage{amsthm}
 \usepackage{mathrsfs}
 \usepackage{float} 
 \usepackage{multirow}  
 \usepackage{xcolor}
 \usepackage[table]{xcolor}
 \usepackage{multirow}
 \usepackage{graphicx}
 \usepackage{booktabs}

\title{KeenKT: Knowledge Mastery-State Disambiguation for Knowledge Tracing}

\author {
    Zhifei Li\textsuperscript{\rm 1,\rm 4,\rm 5},
    Lifan Chen\textsuperscript{\rm 1},
    Jiali Yi\textsuperscript{\rm 1}, 
    Xiaoju Hou\textsuperscript{\rm 2,}\thanks{Corresponding Authors.}, \\
    Yue Zhao\textsuperscript{\rm 3},
    Wenxin Huang\textsuperscript{\rm 1,}\footnotemark[1],
    Miao Zhang\textsuperscript{\rm 1}, 
    Kui Xiao\textsuperscript{\rm 1}, 
    Bing Yang\textsuperscript{\rm 1}
}
\affiliations {
    \textsuperscript{\rm 1}School of Computer Science, Hubei University, Wuhan 430062, China\\
    \textsuperscript{\rm 2}Institute of Vocational Education, Guangdong Industry Polytechnic University, Guangzhou 510300, China\\
    \textsuperscript{\rm 3}Shandong Police College, Ji’nan 250200, China\\
    \textsuperscript{\rm 4}Hubei Key Laboratory of Big Data Intelligent Analysis and Application (Hubei University), Wuhan 430062, China\\
    \textsuperscript{\rm 5}Key Laboratory of Intelligent Sensing System and Security (Hubei University), Ministry of Education, Wuhan 430062, China\\
    \{zhifei1993, zhangmiao, xiaokui\}@hubu.edu.cn, 2023030010@gdip.edu.cn, zhaoy@sdpc.edu.cn,\\ wenxinhuang\_wh@163.com, \{lifan0403, jiali0403\}@stu.hubu.edu.cn, yangbing@126.com

}

\usepackage{bibentry}
\begin{document}

\maketitle

\begin{abstract}
Knowledge Tracing (KT) aims to dynamically model a student’s mastery of knowledge concepts based on their historical learning interactions. Most current methods rely on single-point estimates, which cannot distinguish true ability from outburst or carelessness, creating ambiguity in judging mastery. To address this issue, we propose a \underline{K}nowledge Mast\underline{e}ry-Stat\underline{e} Disambiguatio\underline{n} for \underline{K}nowledge \underline{T}racing model (KeenKT), which represents a student’s knowledge state at each interaction using a Normal-Inverse-Gaussian (NIG) distribution, thereby capturing the fluctuations in student learning behaviors. Furthermore, we design an NIG-distance-based attention mechanism to model the dynamic evolution of the knowledge state. In addition, we introduce a diffusion-based denoising reconstruction loss and a distributional contrastive learning loss to enhance the model’s robustness. Extensive experiments on six public datasets demonstrate that KeenKT outperforms SOTA KT models in terms of prediction accuracy and sensitivity to behavioral fluctuations. The proposed method yields the maximum AUC improvement of 5.85\% and the maximum ACC improvement of 6.89\%.
\end{abstract}

\begin{links}
    \link{Code}{https://github.com/HubuKG/KeenKT}
\end{links}

\begin{figure}[!t]
    \centering
    \includegraphics[width=0.9\linewidth]{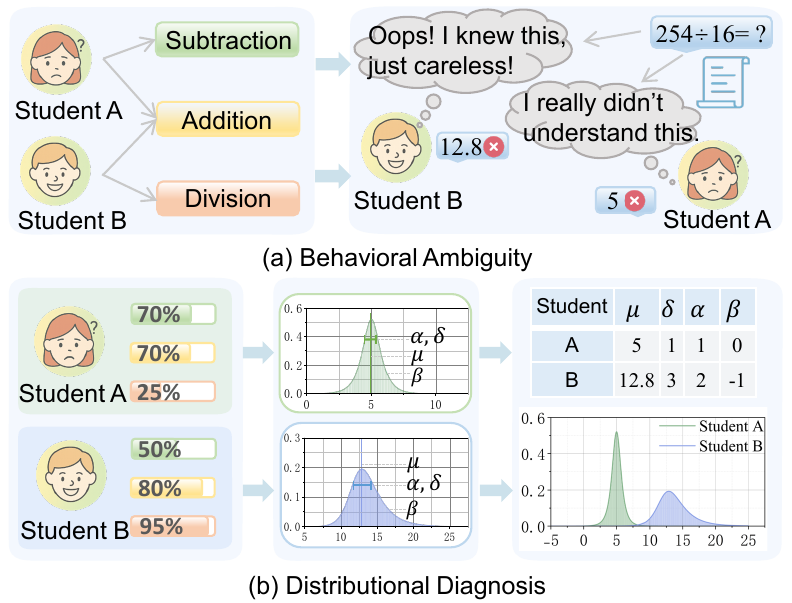}
    \caption{Illustration of behavioral ambiguity in knowledge tracing and our method. (a) Behavioral Ambiguity: The same error was made by two students from different causes (Carelessness and Misunderstanding). (b) Distributional Diagnosis: The Normal-Inverse-Gaussian distribution is used to model the students’ knowledge mastery state.}
    \label{fig1}
\end{figure}
\section{Introduction}

Knowledge Tracing (KT) \cite{survey0} is a foundational technique for enabling personalized education, aiming to dynamically infer a student’s mastery of Knowledge Concepts (KCs) based on their historical learning interactions, and predict future performance \cite{early}. In recent years, the growing scale of learning data and the increasing complexity of student behavior \cite{AdaptKT} have imposed higher demands on the representational power and robustness of KT models.

KT has progressed through three primary phases \cite{survey1}. Initially, early probabilistic models \cite{BKT} represented a student's mastery of each KC as a binary vector, with subsequent models incorporating temporal dependencies \cite{DBKT}. This was followed by the emergence of logistic regression models \cite{ LFA, PFA}, which leveraged exercise-KC interaction features to transform the estimation of mastery probability into interpretable linear or nonlinear relationships. Finally, with the rise of deep learning, deep KT models \cite{DKT, DKT+, DKT-Forget} emerged, utilizing LSTM networks \cite{LSTM} to capture long-range dependencies. Subsequent memory-augmented models \cite{DKVMN, SKVMN} employed external memory mechanisms to store and update knowledge states (KS). Later, attention-based models \cite{SAKT, AKT, SAINT} proved highly effective in capturing global dependencies within sequences. Most recently, graph neural network-based models \cite{GKT, CRKT} have enabled the explicit modeling of relationships between concepts through graph-structured representations.

Despite the notable progress in prediction accuracy, existing methods often struggle to handle sporadic behaviors commonly observed in real learning scenarios, such as outbursts or carelessness. These non-knowledge-related factors are often misinterpreted as fluctuations in student ability, leading to ambiguities in KS estimation. As illustrated in Figure \ref{fig1}(a), \texttt{Student A} and \texttt{Student B} both incorrectly answer the same division question, yet the underlying causes differ substantially: \texttt{Student A} lacks conceptual understanding, while \texttt{Student B}, who has mastered the concept, errs due to carelessness. In such cases, point-estimate-based models \cite{model} cannot distinguish between different causes of identical observable behaviors. To address this limitation, recent studies have explored distributional modeling methods, which represent KS as symmetric probability distributions \cite{UKT}. This allows models to capture the volatility in KS \cite{SparseKT, FlucKT}, but the symmetric distribution can only assume an equilibrium of the distribution, leading to the misestimation of marginalized groups by the model.

\begin{figure*} [!t]
    \centering
    \includegraphics[width=0.9\linewidth]{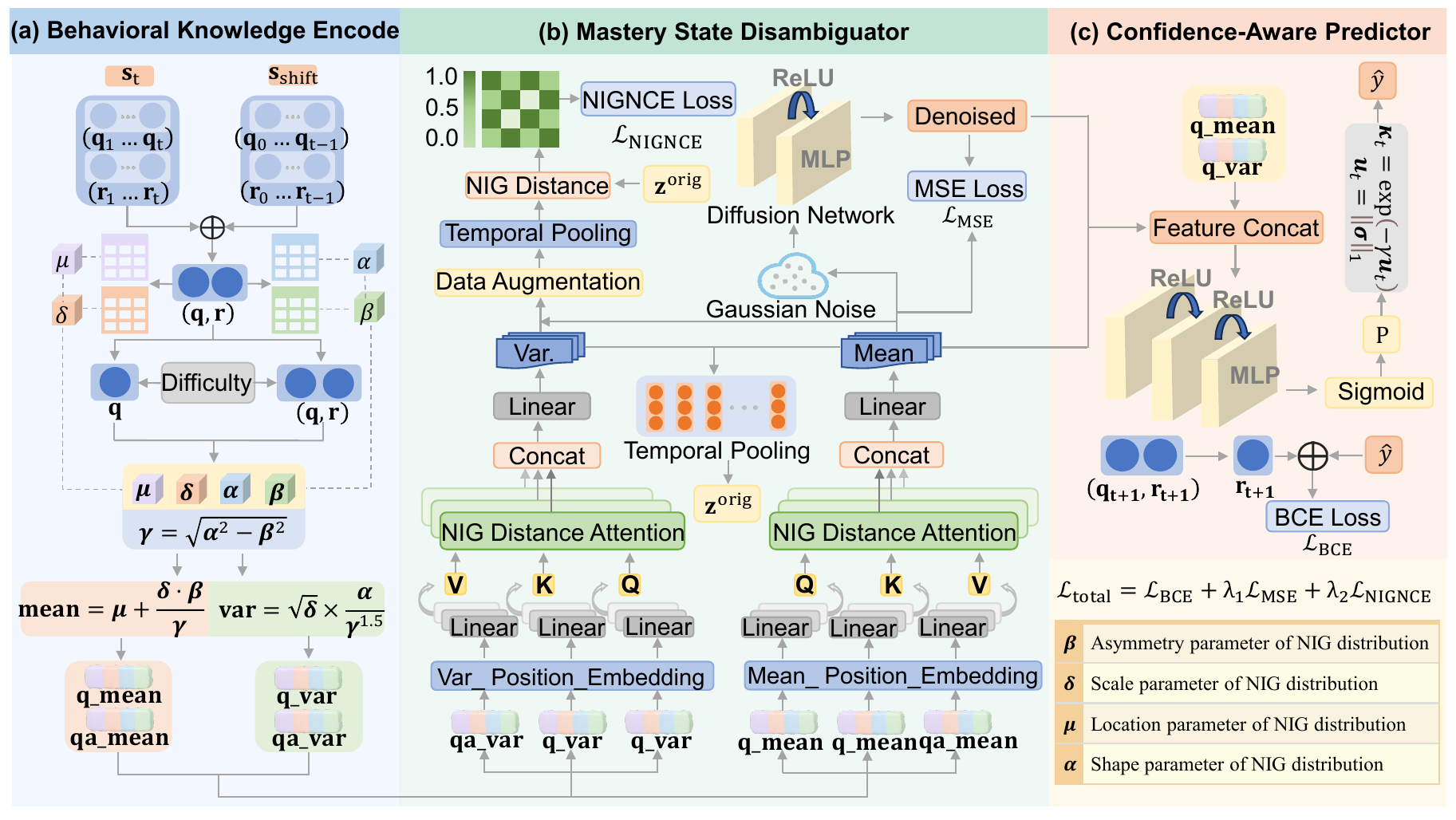}
     \caption{The framework of KeenKT. (a) Behavioral Knowledge Encoder: Models students’ historical interaction sequences and generates distributional mastery representations that capture volatility through a NIG parameterization. (b) Mastery State Disambiguator: Models the temporal evolution of mastery states, incorporating diffusion-based denoising and distributional contrastive learning to enhance robustness. (c) Confidence-Aware Predictor: To estimate the probability of correct responses.}
    \label{fig2}
\end{figure*}
To this end, we propose KeenKT, a KT model designed to disambiguate KS. KeenKT adopts the Transformer-based architecture and represents each student interaction using the asymmetric Normal-Inverse-Gaussian (NIG) distribution, which explicitly encodes the distributional characteristics of mastery states. To further enhance robustness, the model integrates the diffusion-based denoising reconstruction module and the distributional contrastive learning strategy. As shown in Figure \ref{fig1}(b), KeenKT successfully models the KS of \texttt{Students A} and \texttt{Student B} using four-parameter NIG distributions, effectively distinguishing true mastery from behavioral noise based on their respective distributional forms. Our main contributions are as follows:

\begin{itemize}
    \item We introduce the distributional modeling framework for KT by representing students' KS as NIG distributions, enabling the modeling of students' behavioral fluctuation and improving the model’s ability to identify atypical learning behaviors.
    \item We enhance the standard Transformer architecture by replacing the dot-product attention mechanism with a NIG distance-based attention mechanism, and incorporating both diffusion-based denoising reconstruction and distributional contrastive learning to improve robustness.
    \item We conduct extensive experiments on six public benchmark datasets, where KeenKT outperforms 10 competitive baselines in terms of AUC, achieving an average AUC improvement of 2.42\% across six datasets.
\end{itemize}

\section{Related Work}
Knowledge Tracing (KT) is a core task in intelligent educational systems, which aims to dynamically model a student’s mastery of Knowledge Concepts (KCs) based on their historical interactions with learning content. KT methods have evolved from early probabilistic models \cite{BYS} to deep learning-based methods \cite{surveyofdeep}, greatly enhancing their capacity to predict. KT methods can be broadly divided into four categories:

\noindent\textbf{Probabilistic Models}\quad BKT \cite{BKT} is one of the earliest methods for modeling the probability that a student has mastered a specific concept. It treats each concept as a binary latent variable and employs a Hidden Markov Model to represent transitions between the “unmastered” and “mastered” states over time. Based on BKT, FBKT \cite{FBKT} introduces a forgetting parameter, allowing students to regress from a mastered state to an unmastered one, which better reflects realistic cognitive. These models, though powerful, still rely on binary assumptions, paving the way for logistic models.

\noindent\textbf{Logistic Models}\quad LFA \cite{LFA} is a logistic regression–based method that estimates students' knowledge mastery by jointly modeling three key factors: student ability, concept difficulty, and cumulative practice frequency. Building upon this framework, PFA \cite{PFA} further distinguishes the effects of correct and incorrect practice by separately incorporating the number of successful and failed attempts, thereby enhancing the model’s capacity to capture students’ real-time learning states. As datasets expanded, the need for automatic feature learning encouraged a transition from logistic models to deep learning models.

\noindent\textbf{Deep Learning Models}\quad  DKT \cite{DKT} employs LSTM networks \cite{QIKT, LefoKT} to capture long-range temporal dependencies in student learning sequences. DKVMN \cite{DKVMN} and SKVMN \cite{SKVMN} enhance this framework by introducing external memory mechanisms \cite{KQN} to store and dynamically update representations of KCs. To improve the modeling performance of long sequences, the attention mechanism \cite{FoLiBiKT, extraKT} is widely applied. Subsequently, SAKT \cite{SAKT} integrates the Transformer architecture \cite{RKT, HawkesKT, DIMKT, DTransformer, csKT} into the KT domain, capturing interactions from historical responses. GKT \cite{GKT} was the first to introduce Graph Neural Networks (GNNs) into KT, organizing knowledge concepts into a graph structure to model inter-concept relations. These deep models are sensitive to noise, motivating research on robustness‑enhanced models. 

\noindent\textbf{Robustness-Enhanced Models}\quad  Recent studies have proposed various methods to enhance the robustness and structural expressiveness \cite{ IEKT, LPKT, ATDKT, stableKT} of KT models. ATKT \cite{ATKT} introduces an adversarial training mechanism, generating perturbations during training to guide the model toward stable representations under anomalous behavior. UKT \cite{UKT} models the student’s knowledge state as a Gaussian distribution, parameterized by both mean and variance, thereby explicitly capturing uncertainty. 

Existing models predominantly represent students' knowledge states as latent random variables but remain inherently unobservable. This formulation inadequately captures the occasional fluctuations of knowledge states, consequently compromising prediction accuracy.

\section{Methodology}
 In this section, we will introduce KeenKT, as shown in Figure \ref{fig2}. The framework comprises three core components: Behavioral Knowledge Encoder, Mastery State Disambiguator, and Confidence-Aware Predictor.
\subsection{Problem Formulation}
Knowledge Tracing (KT) aims to dynamically model a student's mastery level over latent knowledge concepts (KCs) based on their historical interactions with learning content. Specifically, given a student's past interaction sequence $\mathbf{S}_t=\{(\mathbf{q}_1,\mathbf{r}_1),(\mathbf{q}_2,\mathbf{r}_2),\ldots,(\mathbf{q}_t,\mathbf{r}_t)\}$, where $\mathbf{q}_{i}$ denotes the question at the $i-th$ interaction, and $\mathbf{r}_i\in\{0,1\}$ represents the correctness label of the student’s response. The goal is to predict the probability that the student will answer the next question $\mathbf{q}_{t+1}$ correctly:
\begin{equation}
\hat{y}_{t+1}=\mathrm{P}\left(\mathbf{r}_{t+1}=1\mid \mathbf{S}_{t},\mathbf{q}_{t+1}\right),
\end{equation}
here, $\hat{y}_{t+1}$ is the probability produced by the model, whereas $\mathbf{r}_{t+1}$ is the ground-truth label.
\subsection{Behavioral Knowledge Encoder}
\begin{figure}[!t]
    \centering
    \includegraphics[width=0.9\linewidth]{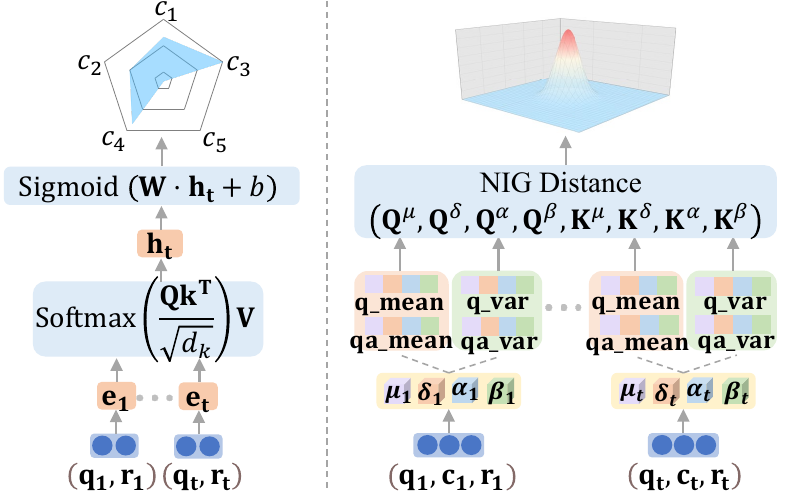}
    \caption{Comparison of sequential modeling. The \texttt{left} panel illustrates traditional KT architectures. The \texttt{right} panel depicts the KeenKT.}
    \label{fig3}
\end{figure}

Traditional KT models represent student states using deterministic vectors \cite{Gotodeepkt}, as shown in Figure \ref{fig3} \texttt{left}, which fail to capture fluctuations. To address this, we propose Behavioral Knowledge Encoder (BKE), which is designed to construct Normal-Inverse-Gaussian (NIG) distribution embeddings, as shown in Figure \ref{fig3} \texttt{right}, by distribution to characterize the student’s knowledge state.

At each time step $t$, in Figure \ref{fig2}(a), BKE concatenates two input sequences for the encoder's final input representation $(\mathbf{q},\mathbf{r})$. Each input is then mapped into four latent vectors by table lookup operations, which correspond to the four parameters of the NIG distribution: $\mu$, $\delta$, $\alpha$, and $\beta$. The parameters are entered into embedding layers and transformed to satisfy distributional constraints:
 \begin{equation}
\left\{
\begin{aligned} 
    \boldsymbol{\mu} &= \mathrm{Embedding}(\mathbf{q}) \\ 
 \boldsymbol{\alpha} &= \mathrm{softplus}(\mathrm{Embedding}(\mathbf{q}))+\varepsilon \\
 \boldsymbol{\beta}  &= \tanh(\mathrm{Embedding}(\mathbf{q}))\cdot\alpha\cdot0.999 \\
 \boldsymbol{\delta} &= \mathrm{ELU}(\mathrm{Embedding}(\mathbf{q}))+1
\end{aligned}
\right.,
\end{equation}where $\varepsilon=10^{-7}$ ensures numerical stability, the scaling by 0.999 constrains $\boldsymbol{\beta}\in(-\boldsymbol{\alpha},\boldsymbol{\alpha}).$ And $\boldsymbol{\delta}$ is enforced to be positive via ELU \cite{ELU} activation. Parameters are then converted into the mean and variance based on the statistical properties of the NIG distribution:
\begin{equation}
\left\{
\begin{aligned} 
\mathbf{mean}&=\boldsymbol{\mu}+\frac{\boldsymbol{\delta}\cdot\boldsymbol{\beta}}{\sqrt{\boldsymbol{\alpha}^2-\boldsymbol{\beta}^2}}\\
\mathbf{var} &=\sqrt{\boldsymbol{\delta}}\cdot\frac{\boldsymbol{\alpha}}{\left(\sqrt{\boldsymbol{\alpha}^2-\boldsymbol{\beta}^2}\right)^{1.5}}
\end{aligned}
\right..
\end{equation}

The encoder ultimately outputs question-level and question–answer representations. 

\subsection{Mastery State Disambiguator}
In KT tasks, student mastery shifts continuously, but outbursts and carelessness add noise that blurs predictions. We introduce the Mastery State Disambiguator (MSD), in Figure \ref{fig2}(b), which couples NIG-distance attention, diffusion-based denoising, and distributional contrastive learning.\medskip

\noindent\textbf{NIG Attention}\quad The module ingests mean- and variance-vector streams from BKE, then sends them into mean and variance pathways, and feeds the results into Transformer \cite{transfomer} for temporal interaction. First, three independent linear projections are applied to derive the mean and variance for Q, K, and V representations, respectively:
\begin{equation}
\left\{
\begin{aligned} \mathbf{Q}_{t}^{\mu}=\mathbf{W}_{Q}^{\mu}\boldsymbol{\mu}_{t};\quad\mathbf{K}_{t}^{\mu}=\mathbf{W}_{K}^{\mu}\boldsymbol{\mu}_{t};\quad\mathbf{V}_{t}^{\mu}=\mathbf{W}_{V}^{\mu}\boldsymbol{\mu}_{t}\\\mathbf{Q}_{t}^{\sigma}=\mathbf{W}_{Q}^{\sigma}\boldsymbol{\sigma}_{t};\quad\mathbf{K}_{t}^{\sigma}=\mathbf{W}_{K}^{\sigma}\boldsymbol{\sigma}_{t};\quad\mathbf{V}_{t}^{\sigma}=\mathbf{W}_{V}^{\sigma}\boldsymbol{\sigma}_{t}
\end{aligned}
\right. \,\,,
\end{equation}
here, $\mathbf{W}$ is learnable parameter matrix, $\mathbf{\mu}$ and $\mathbf{\sigma}$ denote the mean and variance vectors, respectively.  Then, the two vectors are concatenated respectively to obtain a unified triplet:
\begin{equation}
\mathbf{Q}_t=[\mathbf{Q}_t^\mu\|\mathbf{Q}_t^\sigma],\,\mathbf{K}_t=[\mathbf{K}_t^\mu\|\mathbf{K}_t^\sigma],\,\mathbf{V}_t=[\mathbf{V}_t^\mu\|\mathbf{V}_t^\sigma]\in\mathbb{R}^{2d}.
\end{equation}

To capture mastery variation across different time steps, we replace the standard dot-product attention with a similarity measure based on the NIG distribution distance. Specifically, for any two time steps $i$ and $j$, the attention distance is defined as:
\begin{align}
\mathrm{Dist}_{i,j}=\left\|\boldsymbol{\mu}_i-\boldsymbol{\mu}_j\right\|_2^2+\left\|\sqrt{\boldsymbol{\sigma}_i}-\sqrt{\boldsymbol{\sigma}_j}\right\|_2^2.
\end{align}

Then we use an inverse function to get a similarity score:
\begin{align}
s_{ij}=\frac{1}{1+\mathrm{Dist}_{i,j}}.
\end{align}

Followed by temperature-scaled softmax normalization:
\begin{align}
\alpha_{ij}=\frac{\exp(s_{ij}/\tau)}{\sum_k\exp(s_{ik}/\tau)},
\end{align}
where $\tau=0.07$ controls the sharpness of the weight distribution. The resulting sequence:
\begin{align}
{\mathbf{h}}=\sum_j\alpha_{ij}\mathbf{V}_j.
\end{align}
%Representations are aggregated via temporal pooling to form a unified representation $z_{\mathrm{orig}}$, which is further processed by the contrastive learning module.

\noindent\textbf{Diffusion-Based Denoising reconstruction}\quad It introduces controlled noise into the encoded hidden states and attempts to reconstruct the original inputs. The mean squared error (MSE) between the denoised output and the clean input serves as the training objective:
\begin{align}
\mathcal{L}_{\mathrm{MSE}}=\mathrm{MSE}(\mathcal{F}_{\mathrm{diff}}({\mathbf{h}}+\boldsymbol{\varepsilon}),{\mathbf{h}}),
\end{align}
where $\mathcal{F}_{\mathrm{diff}}$ is the denoising function and $\varepsilon$ denotes the injected noise. This objective encourages the model to learn robust representations that are invariant to perturbations.\medskip

\begin{table}[!t]
\centering
\renewcommand{\arraystretch}{1.1}
\small % 使用较小的字体
\setlength{\tabcolsep}{3pt} % 减少列间距
\begin{tabular}{ccccc}
\hline
\multirow{2}{*}{Dataset} & \multicolumn{4}{c}{After Preprocessing} \\ \cline{2-5} 
                         & Interactions & Sequences & Questions & KCs \\ \hline
ASSISTments2009          & 987,606      & 17,056    & -         & 2,748 \\ 
Algebra2005              & 682,797      & 19,093    & -         & 100   \\ 
Bridge2006               & 607,025      & 574       & 173,113   & 112   \\ 
NIPS34                   & 1,817,474    & 1,145     & 129,263   & 493   \\ 
ASSISTments2015          & 1,382,727    & 4,918     & 948       & 57    \\ 
POJ                      & 282,619      & 3,852     & 17,737    & 123   \\ \hline
\end{tabular}
\caption{Statistics of all datasets.}
\label{tab:my-table}
\end{table}

\noindent\textbf{Distributional Contrastive Learning}\quad It treats the original and perturbed sequences as positive pairs and other samples as negatives. A similarity matrix is constructed in distribution space, and the following NIG-based contrastive loss is computed:
\begin{align}
\mathcal{L}_{\mathrm{NIGNCE}}=\mathrm{CE}\left(\begin{bmatrix}-\frac{1}{1+\mathrm{Dist_{ij}^{neg}}},\frac{1}{1+\mathrm{Dist_{ij}^{pos}}}\end{bmatrix}/\tau\right),
\end{align}
where $\mathrm{Dist_{ij}^{\mathrm{neg}}}$ and $\mathrm{Dist_{ij}^{\mathrm{pos}}}$ denote NIG-based distances between negative and positive distribution pairs, respectively. This contrastive objective guides the model to maintain stable prediction capabilities even under behavioral noise.

\begin{table*}[]
\centering
\renewcommand{\arraystretch}{1.06} 
\setlength\tabcolsep{4pt}
\resizebox{\textwidth}{!}{%
\begin{tabular}{c|c|c|cccccc}
\toprule 
\textbf{Models} & \textbf{Venue} & \textbf{Backbone}             & \textbf{AS2009}                             & \textbf{AL2005}                             & \textbf{BD2006}                             & \textbf{NIPS34}                             & \textbf{AS2015}                             & \textbf{POJ}                                \\  \midrule
DKT             & NIPS’15        &                               & 0.8226±0.0011                               & 0.8149±0.0011                               & 0.8015±0.0008                               & 0.7689±0.0002                               & 0.7271±0.0005                               & 0.6089±0.0009                               \\
ATKT            & MM’21          & \multirow{-2}{*}{RNN}         & 0.7470±0.0008                               & 0.7995±0.0023                               & 0.7889±0.0008                               & 0.7665±0.0001                               & 0.7245±0.0007                               & 0.6075±0.0012                               \\ \midrule
GKT             & ICLR’19        &                               & 0.8102±0.0003                               & 0.8154±0.0016                               & 0.8157±0.0005                               & 0.7804±0.0003                               & 0.7101±0.0013                               & 0.6295±0.0007                               \\
DYGKT           & KDD’24         &                               & 0.8168±0.0005                               & 0.8589±0.0008                               & 0.7721±0.0009                               & 0.7704±0.0002                               & 0.7114±0.0009                               & 0.6054±0.0003                               \\
DGEKT           & TOIS’24        & \multirow{-3}{*}{GNN}         & 0.8065±0.0008                               & 0.8769±0.0005                               & 0.7846±0.0010                               & 0.7902±0.0017                               & 0.7153±0.0007                               & 0.6249±0.0004                               \\ \midrule
simpleKT        & ICLR’23        & MLP                           & 0.8413±0.0018                               & 0.9267±0.0003                               & 0.8141±0.0006                               & 0.7966±0.0000                               & 0.7237±0.0005                               & 0.6194±0.0005                               \\ \midrule
SAKT            & EDM’19         &                               & 0.7746±0.0017                               & 0.8780±0.0063                               & 0.7740±0.0008                               & 0.7517±0.0005                               & 0.7114±0.0003                               & 0.6095±0.0013                               \\
AKT             & KDD’20         &                               & 0.8474±0.0017                               & 0.9294±0.0019                               & 0.8167±0.0007                               & 0.7960±0.0003                               & 0.7282±0.0004                               & 0.6218±0.0013                               \\
RobustKT        & WWW’25         &                               & 0.7160±0.0008                               & 0.8205±0.0007                               & 0.8102±0.0004                               & 0.7854±0.0017                               & \cellcolor[HTML]{E8E8E8}{\underline 0.7805±0.0007} & \cellcolor[HTML]{E8E8E8}{\underline 0.6902±0.0003} \\
UKT             & AAAI’25        & \multirow{-4}{*}{Transformer} & \cellcolor[HTML]{E8E8E8}{\underline 0.8563±0.0014} & \cellcolor[HTML]{E8E8E8}{\underline 0.9320±0.0012} & \cellcolor[HTML]{E8E8E8}{\underline 0.8178±0.0009} & \cellcolor[HTML]{E8E8E8}{\underline 0.8035±0.0004} & 0.7267±0.0007                               & 0.6301±0.0005                               \\ \midrule 
\rowcolor[HTML]{FCEEEE}KeenKT          & Ours           & Transformer                   & \textbf{0.8606±0.0011}                      & \textbf{0.9346±0.0008}                      & \textbf{0.8265±0.0009}                      & \textbf{0.8162±0.0011}                      & \textbf{0.8214±0.0006}                      & \textbf{0.7306±0.0002}                      \\ \bottomrule
\end{tabular}
}
\caption{AUC performance of KeenKT and all baselines. Red denotes the best performance, gray denotes the second‑best.}
\label{tab:AUC}
\end{table*}

\subsection{Confidence-Aware Predictor}
Confidence-Aware Predictor aims to predict the student's overall knowledge state. As illustrated in Figure \ref{fig2}(c), this module takes two types of inputs: (1) the embedding of the current item; (2) the latent knowledge state representation from the previous module. These two inputs are concatenated and fed into an MLP to obtain the activation sequence:
\begin{equation}s_t=\mathbf{w}_2^\top\phi\left(\mathbf{W}_1\mathbf{h}_t+\mathbf{b}_1\right)+b_2,\end{equation} where $\phi(\cdot)$ denotes ReLU. It is also able to obtain the probability: $p_{t}=\sigma(s_{t})$. Subsequently, the confidence factor is obtained through the variance vector output by the MSD: \begin{equation}\kappa_t=\exp(-\gamma\cdot\ \|{\boldsymbol{\sigma}}_t\|_1),\label{eq:confidence} \end{equation} where $\gamma$ is onfidence coefficient, affecting the confidence factor $\kappa_{t}$. The final output is the predicted probability $\hat{y}_{t}$ of a correct response at the next time step:
\begin{equation}\hat{y}_t=\kappa_tp_t+(1-\kappa_t)\times0.5.\end{equation}

To optimize predictive performance, we employ a binary cross-entropy (BCE) loss to measure the deviation between the predicted probabilities and the truth responses $r_{t+1}$:
\begin{align}
\mathcal{L}_{\mathrm{BCE}}=-\sum_{t=1}^{T}\left[r_{t}\operatorname{log}\hat{y}_{t}+(1-r_{t})\operatorname{log}(1-\hat{y}_{t})\right].
\end{align}

The overall training objective integrates this prediction loss with two auxiliary objectives from the disambiguation module, forming a multi-task loss function:
\begin{align}
\mathcal{L}_{\mathrm{total}}=\mathcal{L}_{\mathrm{BCE}}+\lambda_{1}\mathcal{L}_{\mathrm{MSE}}+\lambda_{2}\mathcal{L}_{\mathrm{NIGNCE}},
\end{align}where $\lambda_{1}$ and $\lambda_{2}$  are hyperparameters that balance the contribution of denoising and contrastive learning losses.

This predictor not only reflects the student’s current mastery level but also enhances the model’s awareness of knowledge state fluctuations and resilience to anomalous behaviors, leading to more stable and reliable predictions.

\section{Experiments}
In this section, we conduct a series of comprehensive experiments on six publicly available datasets to evaluate the effectiveness of the proposed KeenKT model. The experimental results are designed to address the following six research questions (RQs): \begin{itemize}
    \item \textbf{RQ1}: How does the performance of KeenKT compare to existing knowledge tracing (KT) models?
    \item \textbf{RQ2}: How does each core module impact the performance of KeenKT?
    \item \textbf{RQ3}: How do different hyperparameter configurations affect the performance of KeenKT?
    \item \textbf{RQ4}: How does KeenKT correctly evaluate the accuracy of predictions? 
    \item \textbf{RQ5}: How effectively does KeenKT disambiguate the knowledge mastery state?
    \item \textbf{RQ6}: How does KeenKT achieve targeted modeling of different behavioral patterns?
\end{itemize}  

\begin{table*}[!t]
\centering
\renewcommand{\arraystretch}{1.06} 
\setlength\tabcolsep{4pt}
\resizebox{\textwidth}{!}{%
\begin{tabular}{c|c|c|cccccc}
\toprule 
\textbf{Models} & \textbf{Venue} & \textbf{Backbone}             & \textbf{AS2009}                             & \textbf{AL2005}                             & \textbf{BD2006}                             & \textbf{NIPS34}                             & \textbf{AS2015}                             & \textbf{POJ}                                \\  \midrule
DKT      & NIPS’15 &                               & 0.7657±0.0011                                  & 0.8149±0.0011                                  & 0.8015±0.0008                                  & \cellcolor[HTML]{FCEEEE}\textbf{0.7689±0.0002} & 0.7271±0.0005                                  & 0.6089±0.0009                                  \\
ATKT     & MM’21   & \multirow{-2}{*}{RNN}         & 0.7208±0.0009                                  & 0.7998±0.0019                                  & 0.8511±0.0004                                  & 0.6332±0.0023                                  & 0.7494±0.0002                                  & 0.6075±0.0012                                  \\ \midrule
GKT      & ICLR’19 &                               & 0.7601±0.0016                                  & 0.7707±0.0035                                  & 0.7702±0.0023                                  & 0.7301±0.0016                                  & 0.7604±0.0020                                  & 0.6503±0.0005                                  \\
DYGKT    & KDD’24  &                               & \cellcolor[HTML]{E8E8E8}{\underline 0.7868±0.0005}    & 0.8389±0.0008                                  & 0.8231±0.0008                                  & 0.7201±0.0005                                  & 0.7612±0.0006                                  & 0.6302±0.0009                                  \\
DGEKT    & TOIS’24 & \multirow{-3}{*}{GNN}         & 0.7865±0.0008                                  & 0.8469±0.0005                                  & 0.8442±0.0016                                  & 0.7102±0.0014                                  & 0.7454±0.0005                                  & 0.6446±0.0007                                  \\ \midrule
simpleKT & ICLR’23 & MLP                           & 0.7748±0.0012                                  & 0.8510±0.0005                                  & 0.8510±0.0003                                  & 0.7328±0.0001                                  & 0.7506±0.0004                                  & 0.6498±0.0008                                  \\ \midrule
SAKT     & EDM’19  &                               & 0.7063±0.0018                                  & 0.7954±0.0020                                  & 0.8461±0.0005                                  & 0.6879±0.0004                                  & 0.7474±0.0002                                  & 0.6407±0.0035                                  \\
AKT      & KDD’20  &                               & 0.7772±0.0021                                  & 0.8747±0.0011                                  & 0.8516±0.0005                                  & 0.7323±0.0005                                  & 0.7521±0.0005                                  & 0.6449±0.0010                                  \\
RobustKT & WWW’25  &                               & 0.7651±0.0004                                  & 0.7758±0.0010                                  & 0.7757±0.0011                                  & 0.7354±0.0020                                  & \cellcolor[HTML]{E8E8E8}{\underline 0.7658±0.0012}    & \cellcolor[HTML]{E8E8E8}{\underline 0.6556±0.0005}    \\
UKT      & AAAI’25 & \multirow{-4}{*}{Transformer} & 0.7814±0.0017                                  & \cellcolor[HTML]{FCEEEE}\textbf{0.8781±0.0005} & \cellcolor[HTML]{FCEEEE}\textbf{0.8531±0.0006} & 0.7316±0.0004                                  & 0.7497±0.0002                                  & 0.6548±0.0008                                  \\ \midrule
KeenKT   & Ours    & Transformer                   & \cellcolor[HTML]{FCEEEE}\textbf{0.7934±0.0005} & \cellcolor[HTML]{E8E8E8}{\underline 0.8772±0.0003}    & \cellcolor[HTML]{E8E8E8}{\underline 0.8517±0.0008}    & \cellcolor[HTML]{E8E8E8}{\underline 0.7378±0.0011}    & \cellcolor[HTML]{FCEEEE}\textbf{0.7882±0.0001} & \cellcolor[HTML]{FCEEEE}\textbf{0.7008±0.0004} \\ \bottomrule
\end{tabular}%
}
\caption{ACC performance of KeenKT and all baselines. Red denotes the best performance, gray denotes the second‑best.}
\label{ACC}
\end{table*}

\begin{table}[!t]
\centering
\setlength{\tabcolsep}{4pt}
\small
\renewcommand{\arraystretch}{1.2}
\begin{tabular}{c|cccccc}
\toprule
Models & AS2009 & AL2005 & BD2006 \\
\midrule
w/o CL    & 0.8445±0.0028 & 0.9237±0.0012 & 0.8175±0.0011 \\
w/o Diff. & 0.8432±0.0004 & 0.9228±0.0009 & 0.8143±0.0007 \\
w/o NIG   & 0.8427±0.0010 & 0.9218±0.0018 & 0.8125±0.0012 \\
\midrule
KeenKT    & \cellcolor[HTML]{FCEEEE}\textbf{0.8606±0.0011} & \cellcolor[HTML]{FCEEEE}\textbf{0.9346±0.0008} & \cellcolor[HTML]{FCEEEE}\textbf{0.8255±0.0009} \\
\bottomrule
\end{tabular}
\caption{Ablation study on key components of KeenKT.}
\label{tab:RQ2}
\end{table}

\subsection{Experimental Settings}
\noindent\textbf{Datasets}\quad In this study, we evaluate the performance of the proposed KeenKT model on six widely used public datasets for KT, including ASSISTments2009 (AS2009), Algebra2005 (AL2005), Bridge2006, {NIPS34, ASSISTments2015 (AS2015), and POJ. The key statistics of these datasets are summarized in Table \ref{tab:my-table}. 

\noindent\textbf{Baseline Models}\quad To demonstrate the effectiveness of our model, we compare it with ten baseline grouped into four groups: : (1) RNN-based models: DKT \cite{DKT}, ATKT \cite{ATKT}; (2) GNN-based models: GKT \cite{GKT}, DYGKT \cite{DYGKT}, DGEKT \cite{DGEKT}; (3) MLP-based models: simpleKT \cite{simpleKT}; (4) Transformer-based models: SAKT \cite{SAKT}, AKT \cite{AKT}, RobustKT \cite{RobustKT}, UKT \cite{UKT}. \medskip

\noindent\textbf{Parameter Settings}\quad KeenKT's embedding dimension is set to 128, hidden layer dimension to 256, and the batch size to 128. For all experiments, we set the learning rate to 1e-3, which was found to perform best among le-2, le-3, and le-4. We adopt a student-level five-fold cross-validation for six datasets. An early stopping strategy is also adopted, with a patience of 10 epochs. We employ two metrics to evaluate model performance: Area Under the Receiver Operating Characteristic Curve (AUC) and Accuracy (ACC). 

\subsection{Performance Comparison (RQ1)}
As shown in Table \ref{tab:AUC}, KeenKT consistently outperforms existing mainstream models across all six public datasets in terms of AUC, achieving the best overall performance. Specifically, KeenKT achieves relative improvements of 1.58\% and 5.85\% on datasets with more complex behavioral fluctuations, such as NIPS34 and POJ. Moreover, it demonstrates a clear advantage over the remaining four datasets. In terms of ACC, as shown in Table \ref{ACC}, KeenKT maintains leading performance on four datasets. Whether on the data-regulated AS2015 or the more complex POJ, KeenKT's prediction accuracy has been improved compared to advanced models. Specifically, on AS2015, growth rate is 5.24\%, on POJ, growth rate is 5.85\%

\begin{figure}[!]
    \centering    \includegraphics[width=1.0\linewidth]{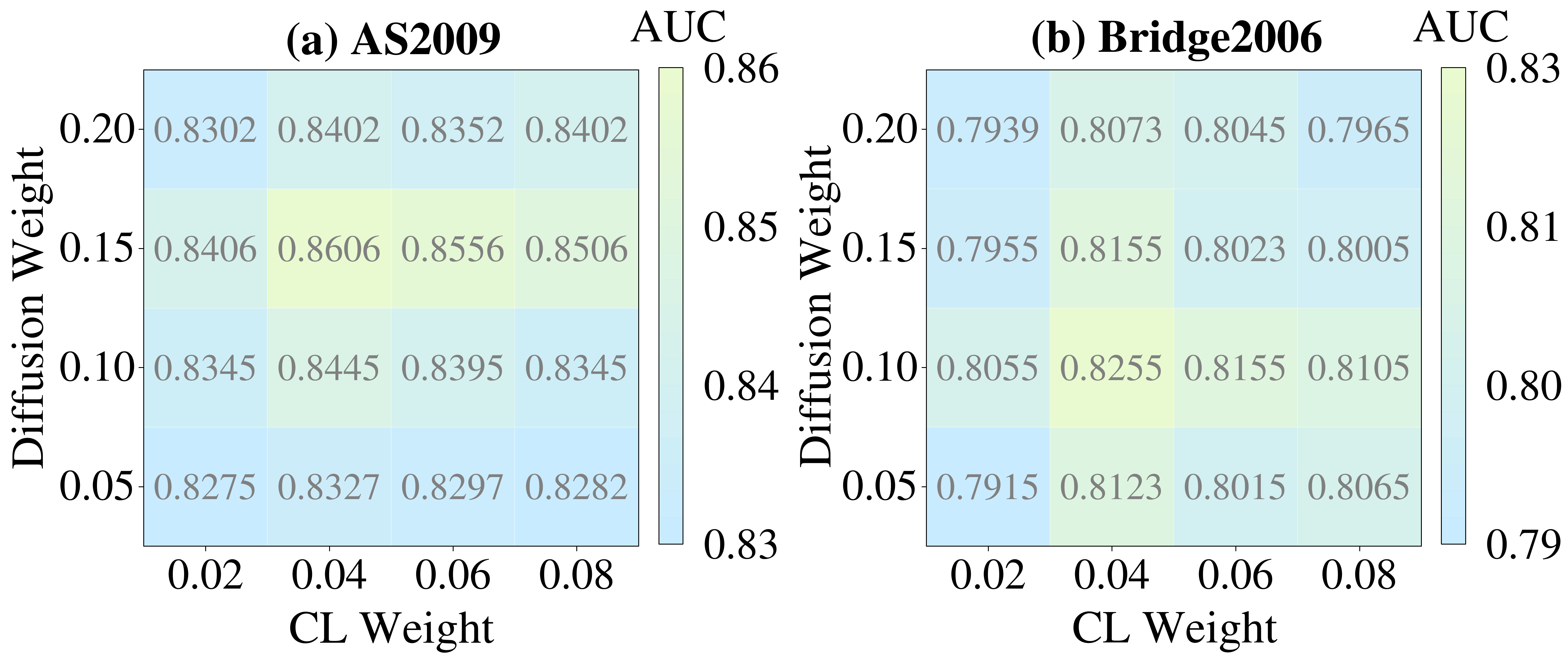}
    \caption{A hyper-parameter analysis on the weight of two modules in (a) AS2009 and (b) Bridge2006.}
    \label{fig4}
\end{figure}

\subsection{Ablation Experiment (RQ2)}
To evaluate the contribution of each key component, we designed three models: (1) w/o CL: removing the distributional contrastive learning module; (2) w/o Diff: removing the diffusion-based denoising reconstruction module; (3) w/o NIG: replacing the NIG distributional embedding with conventional deterministic vector embeddings. 
As shown in Table \ref{tab:RQ2}, eliminating any of the three components degrades performance.
\medskip 

\subsection{Hyperparameter Analysis (RQ3)}
\textbf{Regularization Module Weight Analysis}\quad The weight combination of diffusion denoising reconstruction loss ($\lambda_{1}$) and contrastive learning loss ($\lambda_{2}$) is presented in Figure \ref{fig4}. The optimal results are concentrated around $\lambda_{1}=0.15$ and $\lambda_{2}=0.04$, among which the AS2009 dataset reaches the peak AUC under this configuration. Bridge2006 requires a slightly lower $\lambda_{1}=0.10$. Moderately weighted $\lambda_{1}$ can effectively handle fluctuating behavior, while moderately weighted $\lambda_{2}$ can balance discriminability and stability. Data characteristics determine the optimal weights: tightly clustered distributions need weaker denoising, whereas highly heterogeneous data call for a looser contrastive constraint.\medskip

\begin{figure}[!t]
    \centering
    \hspace*{-0.6cm}
    \includegraphics[width=0.94\linewidth]{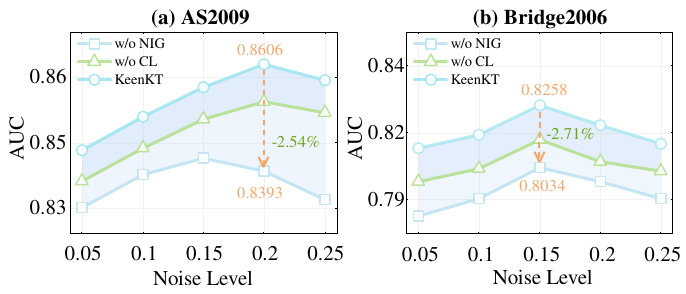}
    \caption{The hyper-parameter analysis on noise level of diffusion-based denoising reconstruction in (a) AS2009 and (b) Bridge2006.}
    \label{fig5}
\end{figure}

\begin{figure}[!t]
    \centering
    \hspace*{-0.18cm}
    \includegraphics[width=1.018 \linewidth]{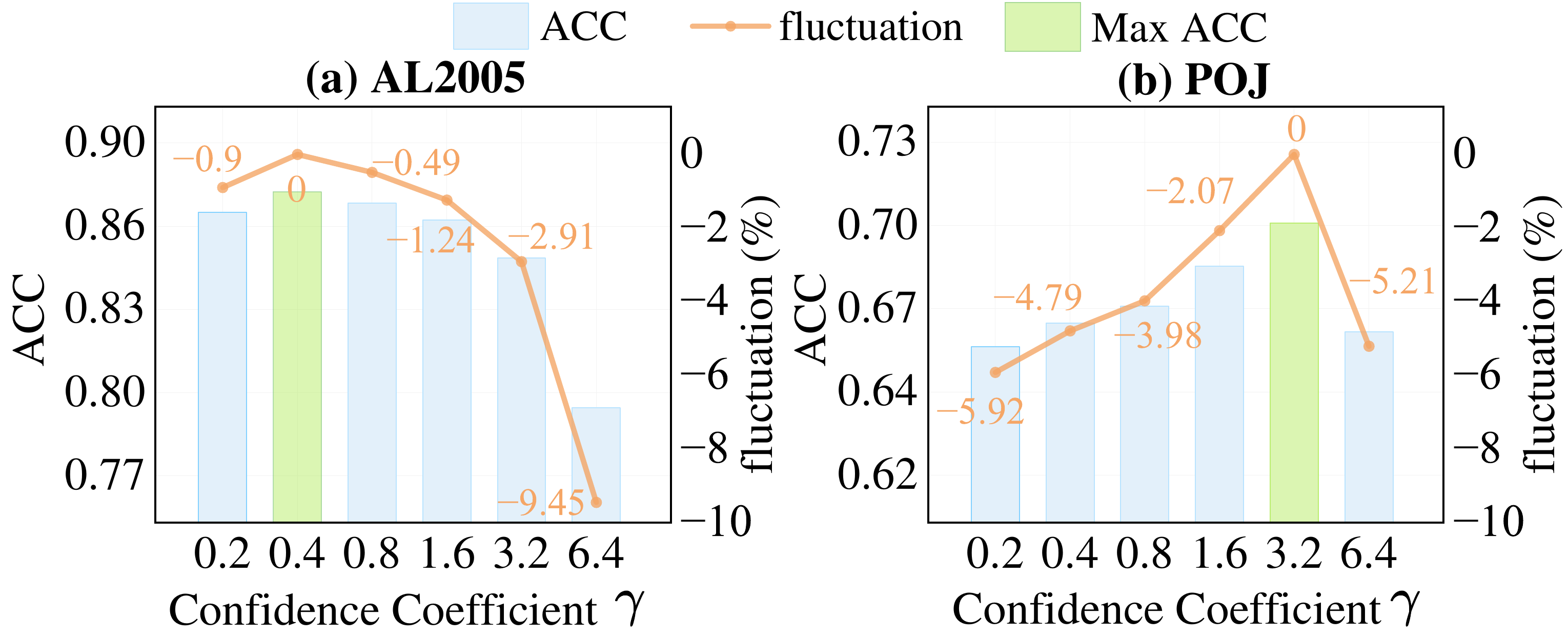}
    \caption{Impact of the confidence coefficient $\gamma$ on accuracy in (a) AL2005 and (b) POJ.}
    \label{fig6}
\end{figure}

\begin{figure*}[!t]
    \centering
    \includegraphics[width=0.96\linewidth]{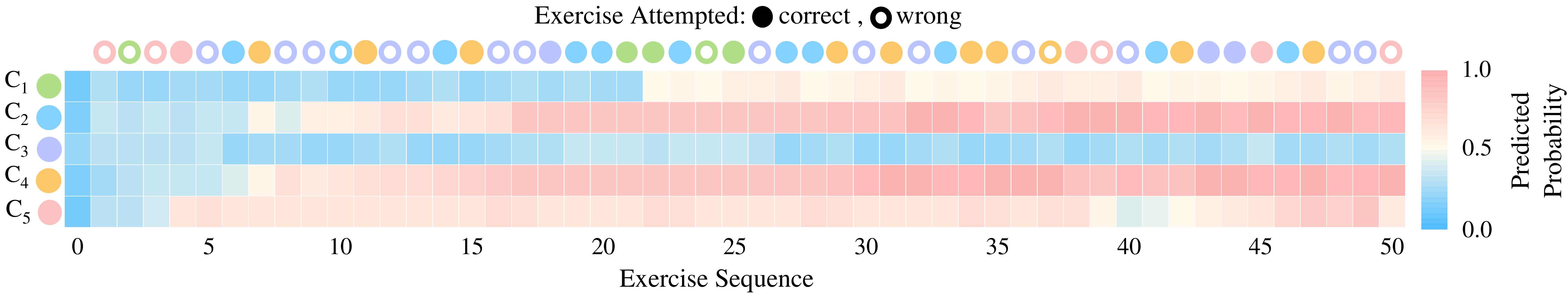}
    \caption{The predicted results of Student \#1 on questions related to five concepts over 50 time steps. All concepts are derived from AS2009. ${C_1}$: Simplifying Expressions; ${C_2}$: Interpreting Algebraic Expressions; ${C_3}$: Evaluating Expressions; ${C_4}$: Ordering Fractions/Decimals; ${C_5}$: Converting Decimals to Fractions.}
    \label{fig7}
\end{figure*}

\begin{figure}[!t]
    \centering
    \includegraphics[width=0.96\linewidth]{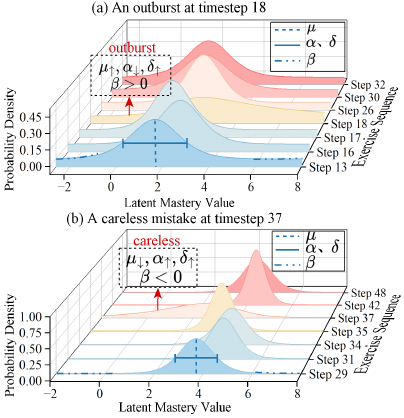}
    \caption{NIG distribution representations of Student \#1 when answering questions related to (a) Concept $C_{3}$ and (b) $C_{4}$ at seven different time steps.}
    \label{fig8}
\end{figure}

\noindent\textbf{Analysis of Diffusion Noise Level}\quad The requirements for noise intensity in different show significant differences. As illustrated in Figure \ref{fig5}. The ASSIST2009 and AL2005 datasets with complex and variable behavioral patterns perform best at strong noise (0.20). The well-structured Bridge2006 requires moderate noise intensity (0.15). Although NIPS34 has high complexity, it reaches its optimum at medium noise intensity (0.15), indicating that excessive perturbation is unnecessary when the dataset is overly complex. These findings indicate that, in highly volatile settings, stronger noise levels facilitate the capture of abrupt shifts. For well-structured data, moderate noise is sufficient to smooth local fluctuations, whereas intrinsically heterogeneous datasets likewise benefit from moderate noise.

\subsection{Confidence Level Analysis (RQ4)}
In KeenKT, the confidence coefficient $\gamma$ controls how heavily the confidence‑aware predictor trusts the NIG variance. As shown in Eq.(\ref{eq:confidence}), larger $\gamma$ magnifies the influence of high‑variance interactions, whereas smaller $\gamma$ down‑weights it. Therefore, we explored the optimal values of gamma on datasets with high volatility (POJ) and low volatility (AL2005). The results are shown in Figure \ref{fig6}. On AL2005, since the overall volatility of the dataset is very small, the model performance reaches its best when the $\gamma$ value is 0.4. On the contrary, on the highly volatile dataset POJ, the model achieves the best performance when the $\gamma$ value is greater than 3.2.

\subsection{Case Study (RQ5)}
To evaluate KeenKT’s capability in capturing abnormal behavioral patterns, we conducted a case study on Student \#1 in the AS2019 dataset. We recorded the model’s predicted mastery probabilities over 50 time steps, as shown in Figure \ref{fig7}. For Concept $C_{4}$, Student \#1 answered correctly at steps 7, 11, 15, 29, and 31, slipped at 37, yet returned to correct answers at 42 and 47; KeenKT’s mastery estimate stayed high despite the single error. For Concept $C_{3}$, the student was wrong at 5, 8, 9, 16, and 17, briefly correct at 18, then wrong again at 26, 30, 32, 36, and 40; the predicted probability did not increase significantly.

This case study demonstrates that KeenKT maintains a coherent representation of knowledge states even in the presence of atypical behaviors, showcasing its advantage in modeling irregular learning trajectories and detecting anomalous patterns.

\subsection{Visual Analysis (RQ6)}
To further investigate KeenKT's modeling capability in knowledge states, we performed a visual analysis of the model’s predicted distributions under two representative interaction scenarios. The results are illustrated in Figure \ref{fig8}.

For burst behaviors, in Figure \ref{fig8}(a), the model exhibits a significant increase in the location parameter $\mu$, indicating improved mastery estimation, while the scale parameter $\delta$ also increases, reflecting reduced confidence. Conversely, when encountering careless errors, in Figure \ref{fig8}(b), KeenKT shows a decrease in $\mu$ under left-skewed $\beta$ values, while $\delta$ increases to capture anomalies. These dynamic adjustments enable precise tracking of students' fluctuating behaviors. 

The observed patterns arise primarily because: (1) The adaptation of $\mu$ accurately reflects short-term variations; (2) Changes in $\alpha$ and $\beta$ distinguish between temporary fluctuations and genuine changes in mastery state; (3) During anomalous events, $\delta$ will automatically adjust to reflect the confidence level.\medskip

\section{Conclusion}
  KeenKT is a novel knowledge tracing framework that combines NIG-distances modeling, a diffusion-based denoising reconstruction network, and a distributional contrastive learning strategy to enhance the ability of discrimination against anomalous behaviors and improve robustness. Future work may extend KeenKT to multimodal learning trajectory modeling, exploring more types of volatility and its potential for real-world deployment in educational settings.

 \section*{Acknowledgments}
This work was supported in part by the National Natural Science Foundation of China (No. 62207011, 62301213, 62407013, 62377009, 62101179), the Natural Science Foundation of Hubei Province of China (No. 2025AFB653), the Natural Science Foundation of Shandong Province of China (No. ZR2024QF257), the Science and Technology Support Plan for Youth Innovation of Colleges and Universities of Shandong Province of China (No. 2023KJ370), the Open Fund of Hubei Key Laboratory of Big Data Intelligent Analysis and Application, Hubei University (No. 2024BDIAA05), and the Open Fund of Key Laboratory of Intelligent Sensing System and Security of Hubei University, Ministry of Education (No. KLISSS202410).

% \bigskip
% \noindent
\bibliography{aaai2026}

\end{document}